\documentclass[10pt,twocolumn,letterpaper]{article}

\usepackage[pagenumbers]{wacv} 

\usepackage{array}
\usepackage{colortbl}
\usepackage{booktabs}
\usepackage{graphicx}
\usepackage{svg}
\usepackage{caption}
\usepackage{multirow} 
\usepackage{makecell}
\usepackage{cuted}
\usepackage{amssymb}
\usepackage{nicefrac}       
\usepackage{xcolor}
\usepackage[table]{xcolor}
\usepackage{tabularx}  
\usepackage{booktabs}
\usepackage[table]{xcolor}
\usepackage{hhline}
\usepackage{stfloats}
\usepackage{wrapfig}

\usepackage{algorithm}
\usepackage{algorithmic}
\usepackage{eqparbox}
\usepackage{amsmath}
\usepackage{amssymb}

\newlength\savewidth

\definecolor{wacvblue}{rgb}{0.21,0.49,0.74}
\usepackage[pagebackref,breaklinks,colorlinks,allcolors=wacvblue]{hyperref}

\def\wacvPaperID{290} 
\def\confName{WACV}
\def\confYear{2026}

\title{Mitigating Object and Action Hallucinations in Multimodal LLMs via Self-Augmented Contrastive Alignment}

\author{
    {Kai-Po Chang}$^{1,\dagger}$, {Wei-Yuan Cheng}$^1$, {Chi-Pin Huang}$^1$, {Fu-En Yang}$^2$, and {Yu-Chiang Frank Wang}$^{1,2,\ddagger}$\\
    \normalsize \textsuperscript{1} Graduate Institute of Communication Engineering, National Taiwan University\\
    \normalsize \textsuperscript{2} NVIDIA\\
    {\tt\small $^{\dagger}$f11942093@ntu.edu.tw, $^{\ddagger}$frankwang@nvidia.com}
}

\begin{document}
\maketitle
\begin{abstract}
\label{sec:abstract}

\vspace{-2mm}
Recent advancement in multimodal LLMs (MLLMs) has demonstrated their remarkable capability to generate descriptive captions for input videos. However, these models suffer from factual inaccuracies in the generated descriptions, causing severe hallucination issues. While prior works have explored alleviating hallucinations for static images, jointly mitigating visual object and temporal action hallucinations for dynamic videos remains a challenging and unsolved task. To tackle this challenge, we propose a \textbf{S}elf-\textbf{A}ugme\textbf{n}ted Con\textbf{t}rastive \textbf{A}lignment (SANTA) framework for enabling object and action faithfulness by exempting the spurious correlations and enforcing the emphasis on visual facts. SANTA employs a hallucinative self-augmentation scheme to identify the potential hallucinations that lie in the MLLM and transform the original captions to the contrasted negatives. Furthermore, we develop a tracklet-phrase contrastive alignment to match the regional objects and relation-guided actions with their corresponding visual and temporal phrases. Extensive experiments demonstrate that SANTA outperforms existing methods in alleviating object and action hallucinations, yielding superior performance on the hallucination examination benchmarks.

\vspace{-3mm}

\end{abstract}
\section{Introduction}
\label{sec:intro}

Recently, Large Language Models (LLMs)~\cite{llama3, qwen2} have demonstrated extensive reasoning capability in NLP tasks and exhibit great potential to be leveraged in vision-language models to handle multimodal understanding tasks, such as visual captioning~\cite{dream1k, msrvtt} and question answering~\cite{videomme,activitynetqa}. However, these multimodal LLMs (MLLMs) are prone to generate descriptions that are not grounded in the given visual input, causing severe \textit{hallucination} issues~\cite{vidhal,factvc,vript,videohallucer}. It would pose substantial risks in real-world applications that require high-standard robustness and faithfulness, like healthcare and autonomous driving. As a result, mitigating hallucinations for multimodal understanding draws significant attention from both academics and industry.


\begin{figure}[t]
  \centering
   \includegraphics[width=1\linewidth]{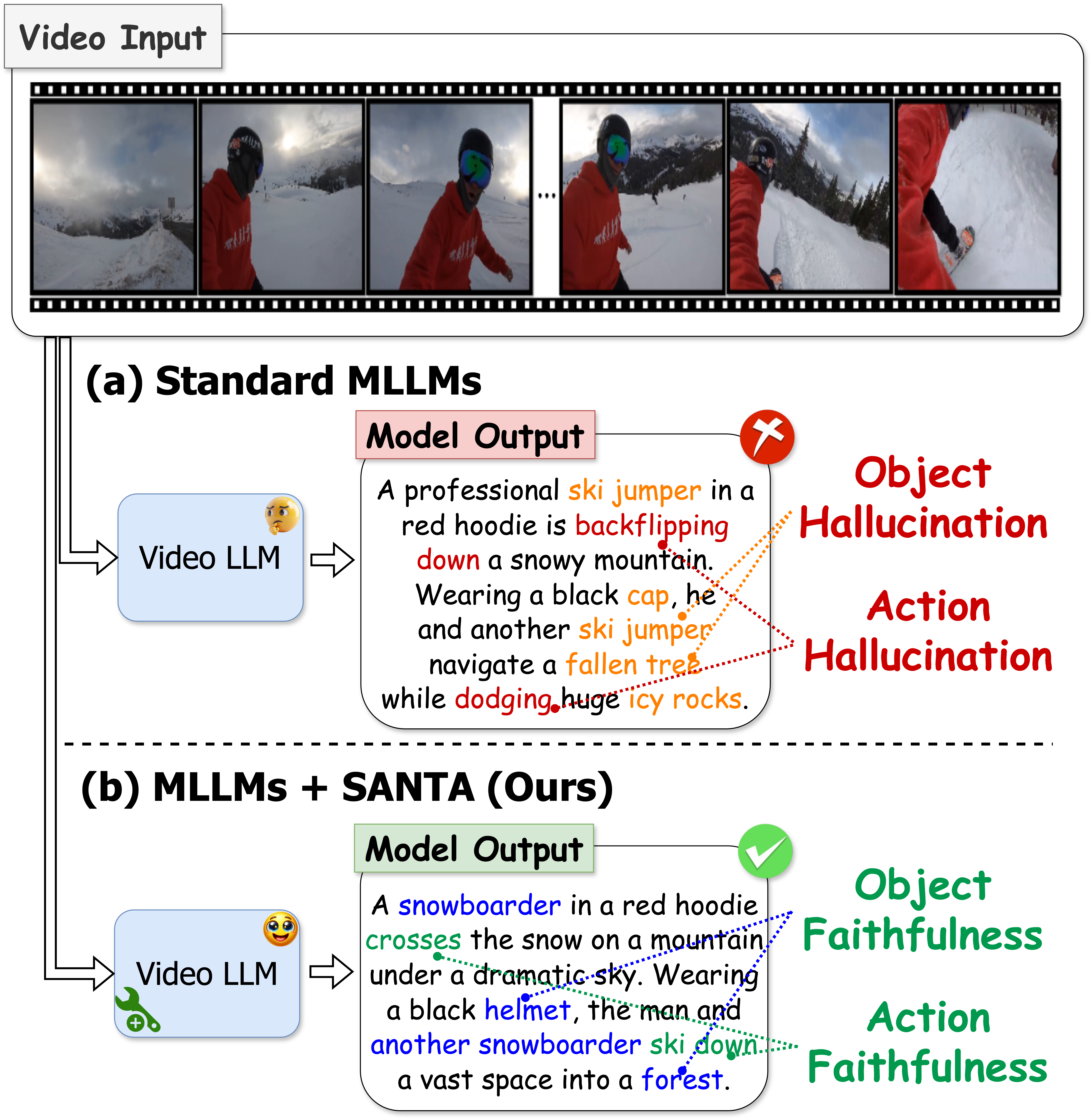}
   \caption{Compared to existing MLLMs suffering from object and action hallucinations, our \textit{SANTA} enhances faithfulness in describing both visual objects and temporal actions.}
   \label{fig:teaser}
   \vspace{-4.5mm}
\end{figure}

To deal with this challenge, several works~\cite{hacl,deco,halva,perturbollava,aip,vcd,opera} have explored mitigating hallucinations in \textit{image} understanding tasks, where the output captions might contain hallucinated \textit{objects} that do not exist in the given inputs. For example, several works~\cite{deco, vcd, opera} enforce that the visual inputs can be sufficiently considered during the language decoding process. However, these methods induce extra computational overhead, substantially slowing down the inference process. On the other hand, HACL~\cite{hacl} and HALVA~\cite{halva} align the visual tokens with textual ones via contrastive learning to encourage the visual information to be interpretable by LLMs. While the object hallucinations are alleviated, the above methods only focus on the hallucinated objects that appear in a static \textit{image} but cannot handle the possible hallucinations of visual objects and temporal actions that commonly occur in \textit{video} domains.

With the goal of mitigating hallucinations of objects and actions, recent approaches~\cite{vistallama,vidhal,vript} have designed learning strategies to capture spatial-temporal dynamics in video data. Vista-LLaMA~\cite{vistallama} introduces an attention mechanism that equalizes the influence of visual and textual tokens on language outputs. Nevertheless, Vista-LLaMA~\cite{vistallama} treats video data in a frame-by-frame manner, leading to the difficulties of MLLMs in capturing the accurate visual appearance of objects and temporal dynamics of actions. Therefore, enabling the output descriptions of MLLMs to be sufficiently faithful in regard to both visual objects and temporal actions remains a challenging yet unsolved problem.


In this paper, we propose a \textit{\textbf{S}elf-\textbf{A}ugme\textbf{n}ted Con\textbf{t}rastive \textbf{A}lignment (SANTA)} framework that mitigates object and action hallucinations occurred in video understanding tasks, as depicted in Fig.~\ref{fig:teaser}. To enforce the output descriptions to be exempted from the spurious correlation caused by language prior while adequately grounding into visual content, \textit{SANTA} employs a~\textit{hallucinative self-augmentation} scheme to enhance a~\textit{tracklet-phrase contrastive alignment}. More specifically, the former observes the phrase-level spurious correlation of the MLLM, and then augments the original captions to serve as the contrasted negatives. The latter first represents visual objects and temporal actions by deriving visual tokens from region-level tracklets and object relations, respectively. With such spatial-temporal visual features obtained, we are able to bridge the modality gap by aligning object and action features with the corresponding language tokens while contrasting the hallucinative ones. By iteratively exploiting the above augmenting and training stages, our \textit{SANTA} enables the alleviation of visual object and temporal action hallucinations in pre-trained MLLMs.  


We now summarize the contributions of this work below:
\begin{itemize}
  \item We present \textit{SANTA}, which jointly mitigates object and action hallucinations by suppressing hallucinative correlation from language prior while enhancing the emphasis of tracklet-level visual facts.
  \item \textit{SANTA} employs a hallucinative self-augmentation scheme to identify potential hallucinations in MLLMs and construct contrasted negatives to simulate this underlying hallucinative correlation.
  \item We develop a tracklet-phrase contrastive alignment to match the regional objects and relation-aware actions with their associated visual and temporal phrases while expelling the hallucinative negatives.
\end{itemize}

\section{Related Works}
\begin{figure*}[t!]
  \centering
  \vspace{-7mm}
  \includegraphics[width=1.0\textwidth]{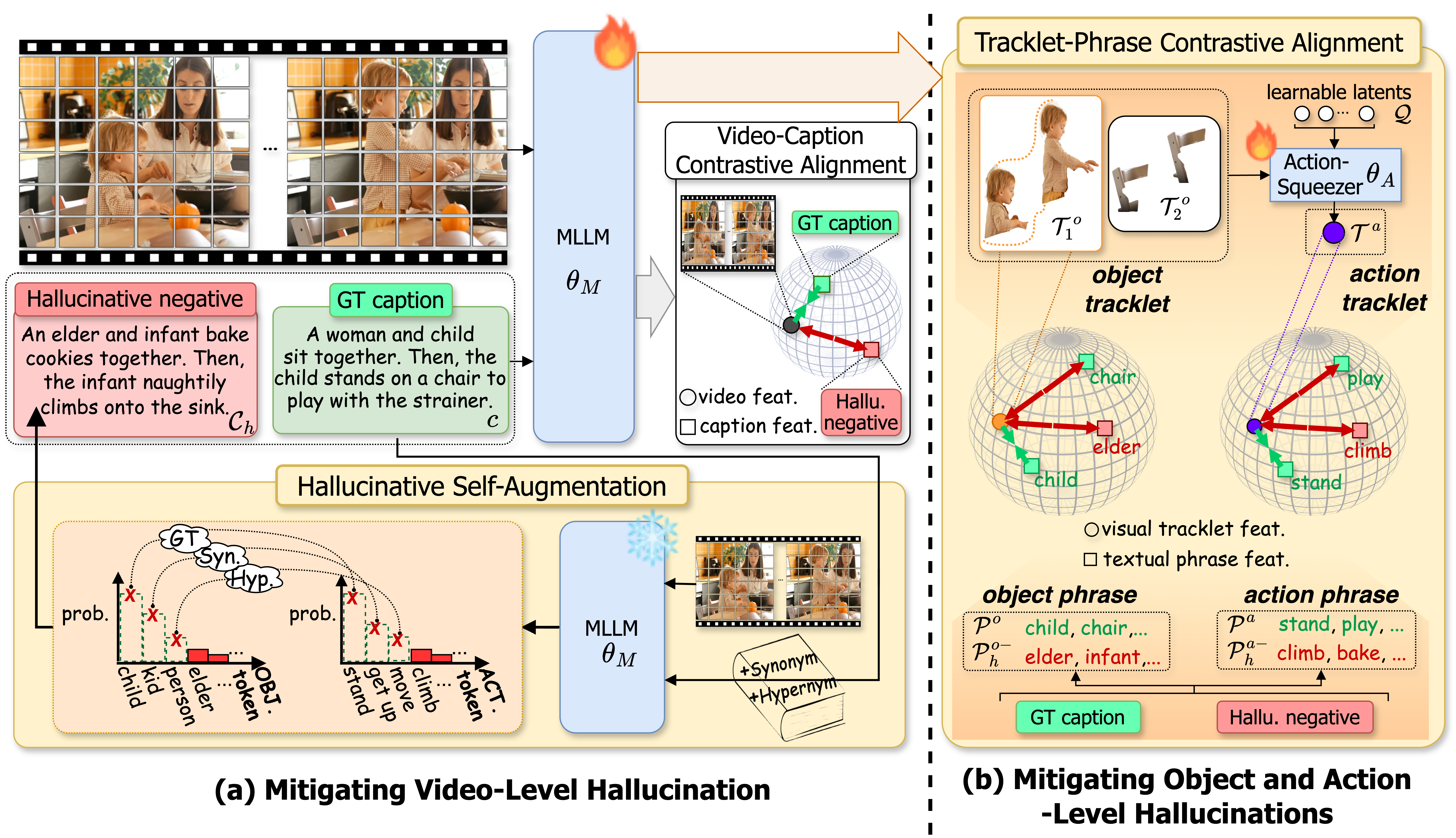}

  \vspace{-2mm}
  \caption{
    \textbf{Overview of \textit{SANTA}.} 
    We employ (a) Mitigating Video-Level Hallucination by applying Hallucinative Self-Augmentation to identify the highly potential hallucinated tokens in MLLM $\theta_M$ that deviate from ground truth words (e.g., synonyms or hypernyms) and then perform video-caption contrastive alignment. \textit{SANTA} then (b) Mitigating Object nad Action-Level Hallucinations by Tracklet-Phrase Contrastive Alignment to align object and action tracklets with visual and temporal phrases while contrasting hallucinative negatives. 
  }
  \vspace{-3mm}
  \label{fig:overview}

\end{figure*}

\subsection{Mitigating Object Hallucination in MLLMs}

In recent years, multimodal LLMs (MLLMs) have demonstrated significant achievement in vision-language tasks (e.g., image captioning~\cite{dci} and visual question answering~\cite{okvqa}). However, MLLMs tend to generate descriptions based on LLM's language prior instead of grounding the facts in the visual condition, causing hallucinated objects that do not exist in the given images. For example, Visual Contrastive Decoding (VCD)~\cite{vcd} reduces hallucination by subtracting language-prior probabilities from the generated token distribution. DeCo~\cite{deco} calibrates token selection by weighting logits from the final and an anchor layer that exhibits the highest confidence in token generation. While promising, the above methods incur additional computational burden during the language decoding process, resulting in a large increase in inference time.


On the other hand, another line of works~\cite{hacl,halva,perturbollava,rapper,ptuning,aip} aims to align visual and textual tokens, so that LLMs could sufficiently interpret and ground the visual facts. For instance, HACL~\cite{hacl} aligns the global image and caption feature to reduce the modality gap. Building upon this, HALVA~\cite{halva} proposes an alignment-based training objective that aligns vision and corresponding textual representations at the phrase level. However, these approaches can only handle the object hallucinations in a static image, limiting the applicability to alleviate the hallucinated temporal actions that are commonly encountered for video data.



\subsection{Mitigating Action Hallucination in MLLMs}

Action hallucination occurs when models describe actions absent from the video, while most existing works instead study relationship hallucination, which focuses on static relationships between objects in a single image. For example, Tri-HE~\cite{tri_he} evaluates relations as edges between object nodes in image scene graphs, while R-Bench~\cite{r_bench} assesses relationship hallucination with image-level and instance-level questions derived from static image captions. These benchmarks focus on evaluation for static images and cannot address the multi-frame reasoning needed to detect action hallucinations. Mitigating action hallucination, in contrast, requires learning to track the evolving states of dynamic objects and to understand cross-frame interactions between multiple dynamic entities in order to infer motion, such as distinguishing a person going up vs. down stairs.

Building on this insight, recent works attempt to encourage captions to be grounded in the spatial-temporal visual content. For example, Vista-LLaMA~\cite{vistallama} introduces a model architecture that ensures a consistent influence of visual tokens on each generated language token. Vriptor~\cite{vript} builds on previous findings that detailed captions improve vision-language alignment~\cite{llava,llavanext} by incorporating voice-over and audio transcriptions. Nevertheless, processing videos frame by frame cannot exploit the temporal dynamics needed for action faithfulness. To tackle this challenge, we employ a \textit{SANTA} framework that jointly aligns object and action tracklets with the corresponding textual tokens, enabling outputs with stronger object and action faithfulness.

\label{sec:formatting}
\section{Method}

\subsection{Problem Formulation}

Given an input video $v$, the goal of the MLLM $\theta_{\textit{M}}$ is to generate a textual caption $c = \{w_i\}_{i=1}^{|c|}$ with length $|c|$, where each token $w_i$ may describe an object, an action, or other contextual information. 
A hallucination occurs when a predicted token $w_i$ refers to a visual concept that is \textit{not} present in the video $v$. Formally, let $\mathcal{O}_v$ and $\mathcal{A}_v$ denote the sets of objects and actions that actually appear in $v$, respectively. We define an \textbf{object hallucination} as the case where $w_i$ is an object noun but $w_i \notin \mathcal{O}_v$, and an \textbf{action hallucination} as the case where $w_i$ is an action verb but $w_i \notin \mathcal{A}_v$.

To mitigate object and action hallucinations, we propose \textit{SANTA}, which jointly suppresses spurious correlations from language priors and enforces proper video grounding. 
Our framework leverages contrastive alignment to identify spurious correlations and construct hallucinative negative samples in a bootstrapped manner (Sec.~\ref{sec:3.2}). 
Furthermore, instead of naively bridging the video-language gap at the frame level, we introduce a novel \textit{tracklet-phrase contrastive learning} that aligns tracklet-level object and action features with their corresponding object and action phrases in the caption (Sec.~\ref{sec:3.3}).



\subsection{Mitigating Video-Level Hallucination}
\label{sec:3.2}

To enforce the LLMs to properly focus on the visual facts rather than the superior correlation inherited in the language prior, we present a contrastive learning framework that aligns the visual features to the token space of LLMs while exempting the potential hallucination descriptions. Instead of directly prompting off-the-shelf LLMs (e.g., GPT-4~\cite{gpt4}) to produce hallucinative captions as hard negative samples, we propose a \textit{Hallucinative Self-Augmentation} strategy to construct the contracted negatives that reflect the inherent spurious correlations of the MLLM $\theta_M$.

\vspace{-2mm}

\paragraph{Hallucinative Self-Augmentation.}

As described in the Fig.~\ref{fig:overview}(a), given a video-caption pair $(v,c)$, we reveal the potential hallucination of the MLLM $\theta_M$ by identifying the text tokens with the highest likelihood that do not belong to the set of ground truth tokens. More specifically, we generate a hallucinative caption by constraining the frozen MLLM $\theta_M$ to select the token with the highest-probability (at each decoding step) that falls \textit{outside} the ground-truth tokens. The set of ground-truth tokens is constructed from the object nouns and action verbs in the ground-truth caption, together with their synonyms and hypernyms (defined by WordNet~\cite{wordnet}). As a result, the hallucinative captions $C_{h}=\{c_i\}_{i=1}^{|C_h|}$ could be generated as follows:
\begin{equation}\label{eq:bootstrapped_hn}
C_h \sim \prod_{t=1}^{T} P_{\theta_M}^{\text{sup}}(c_t \mid v, c_{<t}), \text{where}
\end{equation}
\begin{equation*}
P_{\theta_M}^{\text{sup}}(c_t \mid v, c_{<t}) = \mathbb{I}(c_t) \cdot
\frac{ P_{\theta_M}(c_t \mid v, c_{<t})}
{\sum_{c \notin \Omega} P_{\theta_M}(c \mid v, c_{<t})},
\end{equation*}
where $c_t$ is the generated token at step $t$ , and $\mathbb{I}(\cdot)$ is  an indicator function defined such that $\mathbb{I}(c_t)=0$ if $c_t \in (O \cup A)$ and $\mathbb{I}(c_t)=1$ otherwise. $O$ and $A$ are the set of ground-truth objects and action tokens, including their respective synonyms and hypernyms.

\vspace{-2mm}
\paragraph{Video-Caption Alignment with Hallucinative Negatives.}

With the hallucinative captions obtained, we aim at matching the video features to the LLM's token space and effectively suppressing the possible object and action hallucinations caused by language correlations. Concisely, we impose video-caption contrastive learning to attract the visual feature of input video $v$ with the textual token of the paired caption $c$ while pushing away the negative samples ($V^{-}$ and $C^{-}$) that contain hallucinative caption as hard negatives and randomly sampled videos/captions. Formally, $\mathcal{L}_{\text{video}}$ is defined as follows:

\begin{equation}\label{eq:loss_video}
\small
\mathcal{L}_{\text{video}} = \frac{1}{2} \left[\mathcal{L}_{\text{t2v}}(c,v^{+},V^{-})+ \mathcal{L}_{\text{v2t}}(v,c^{+},C^{-})\right],
\end{equation}
where
\[\label{eq:coarse_c2v_loss}
\small
\mathcal{L}_{\text{t2v}}(c,v^{+},V^{-}) = -\mathbb{E}_{} \left[\log \frac{\phi(c, v^{+})}
{\phi(c, v^{+})+\sum_{i=1}^{B} \phi(c, V^{-}_i)}\right],
\]

\[\label{eq:coarse_v2c_loss}
\small
\mathcal{L}_{\text{v2t}}(v,c^{+},C^{-}) = -\mathbb{E}_{} \left[\log \frac{\phi(v, c^{+})}
{\phi(v, c^{+})+\sum_{i=1}^{B} \phi(v, C^{-}_i)}\right],
\]
where $\phi(\cdot, \cdot) = e^{ \text{sim}(\cdot, \cdot) / {\tau}}$ denotes the cosine similarity between pooled vision and text features. $\tau$ is a temperature parameter. $V^{-}=\{V^{-}_{1},...,V^{-}_{B}\}$ and $C^{-}=\{C^{-}_{1},...,C^{-}_{B}\}$ are the sets of negative in-batch video and caption samples. $B$ is the input batch size.




\subsection{Mitigating Object and Action-Level Hallucinations with Hallucinative Negatives}
\label{sec:3.3}

While video-caption alignment bridges the gap between visual and textual modalities, it remains challenging to precisely locate the object regions and temporal actions in a dynamic video. As illustrated in Fig.~\ref{fig:overview}(b), we introduce a novel \textit{Tracklet-Phrase Contrastive Learning} paradigm to enforce the MLLM to accurately interpret the appeared objects and actions for generating faithful captions.


\vspace{-3.5mm}
\paragraph{\mbox{Enhancing Object-Level Faithfulness.}}

To ensure object faithfulness in the output descriptions of MLLM, we aim to associate object tracklets extracted from the video $v$ with their corresponding object phrases derived from the ground-truth caption $c$. More specifically, we derive the feature of regional object tracklets $\mathcal{T}^{o}$ by employing existing visual grounding methods (e.g., Grounding-SAM2~\cite{groundingdino1.5, sam2}), and obtain the phrase-level textual tokens $\mathcal{P}^{o}$ by parsing with text-only LLMs (e.g., GPT-4~\cite{gpt4}). Subsequently, we define the regional-level object alignment loss $\mathcal{L}_{\text{obj}}$ as:
\begin{equation}\label{eq:loss_obj}
\small
\mathcal{L}_{\text{obj}} = \frac{1}{2}\left[\mathcal{L}_{\text{t2v}}(\mathcal{T}^{o},\mathcal{P}^{o},\mathcal{P}^{o-})+ \mathcal{L}_{\text{v2t}}(\mathcal{P}^{o},\mathcal{T}^{o},\mathcal{T}^{o-} \cup \mathcal{T}^{o-}_h) \right],
\end{equation}
where $\mathcal{P}^{o-}$ and $\mathcal{T}^{o-}$ are the sets of negative in-batch object phrases and tracklets. Note that the object phrases $\mathcal{T}^{o-}_h$ parsed from the hallucinated caption from Eq.~\ref{eq:bootstrapped_hn} serve as hard negatives to augment the region-level object alignment for enhancing object faithfulness.

\vspace{-3mm}
\paragraph{\mbox{Enhancing Action-Level Faithfulness.}}


To extract the temporal actions from a dynamic video, we propose to derive relation-guided action features from visual object tracklets. To achieve this, we design a perceiver-based~\cite{perceiver,blip2,paxion} \textit{action squeezer}  $\theta_A$ to observe the interaction between object tracklets and extract the action tracklet features $\mathcal{T}^{a}$ associated to the action verb $a$ in the ground truth caption. More concisely, $\theta_A$ employs a set of learnable queries $\mathcal{Q}=\{q_k\}_{k=1}^{N_{\mathcal{Q}}}$ to derive $\mathcal{T}^{a}$ from its corresponding set of $N$ related object tracklet features $\{\mathcal{T}^o_i\}_{i=1}^{N}$ specified by the relationships in the ground-truth captions. Through the cross-attention mechanism in $\theta_A$, queries $\mathcal{Q}$ are able to interact and capture the action information lies in the involved object tracklet features $\{\mathcal{T}^o_i\}_{i=1}^{N}$. Then, the action $\mathcal{T}^{a}$ could be selected from the output of $\theta_A$ based on the highest similarity to the corresponding action phrase feature $\mathcal{P}^a$. The above process is formulated as follows,  
\begin{equation}
\begin{split}
    \mathcal{T}^a &= \theta_M(q_{k^*}), \text{where}\\
    k^* &= \operatorname*{argmax}_{k \in \{ 1, \dots, N_{\mathcal{Q}} \}} \Big\{ \text{sim}\Big(\theta_A(q_k,\{\mathcal{T}^o_i\}_{i=1}^{N}),\mathcal{P}^a\Big) \Big\}.
\end{split}
\end{equation}
As a result, the relation-guided action alignment $\mathcal{L}_{\text{act}}$ is computed as:
\begin{equation}\label{eq:loss_act}
\small
\mathcal{L}_{\text{act}} = \frac{1}{2}\left[\mathcal{L}_{\text{t2v}}(\mathcal{T}^{a},\mathcal{P}^{a},\mathcal{P}^{a-})+ \mathcal{L}_{\text{v2t}}(\mathcal{P}^{a},\mathcal{T}^{a},\mathcal{T}^{a-} \cup \mathcal{T}^{a-}_h) \right],
\end{equation}
where $\mathcal{P}^{a-}$ and $\mathcal{T}^{a-}$ are the sets of negative in-batch action phrases and tracklets. 

With our proposed tracklet-phrase contrastive alignment, the spatial-temporal visual tracklet features are effectively captured and aligned with their corresponding language phrase tokens. As a result, the captions generated by MLLMs exhibit sufficient object and action faithfulness.

\subsection{Summary of~\textit{\textbf{SANTA}}}
\begin{algorithm}[t]
\caption{Training of~\textit{SANTA}}
\label{algo}
\begingroup
\setlength{\intextsep}{0pt} 
\setlength{\textfloatsep}{0pt}
\footnotesize 

\textbf{Model:}\\Targeted MLLM $\theta_M$, Action-squeezer $\theta_A$, Learnable queries $\mathcal{Q}$ \\
\textbf{Input:}\\Given batched video-caption pairs $\mathcal{B}=\{v_j, c_j\}_{j=1}^{|B|}$, each caption $c_j$ contains multiple action instances. Each action instance $a$ is associated with multiple object instance $\{o_i\}_{i=1}^N$, where $a,o_i \in c_j$, and the object tracklet features for each $o_i$ are denoted as $\mathcal{T}_i^{o}$.


\begin{algorithmic}[1]
    \FOR{$t=1,\dots,T$}
        \STATE \textcolor{blue}{\textit{/* Hallucinative Self-Augmentation */} }
        \STATE Freeze $\theta_M$;
        \STATE $C_h$ $\leftarrow$ Generate the hallucinative caption by Eq.~\ref{eq:bootstrapped_hn};
        \STATE \textcolor{blue}{\textit{/* Contrastive Alignment */} }
        \STATE $\mathcal{L}_{\text{video}}$ $\leftarrow$ Compute video-caption alignment with $\mathcal{B}$ and $C_h$ by Eq.~\ref{eq:loss_video};
        \STATE $\mathcal{P}^o_i$ $\leftarrow$ Get object phrase features with $\theta_M(o_i)$;
        \STATE $\mathcal{L}_{\text{obj}}$ $\leftarrow$ Compute \textit{Region-Level Object Alignment} with object tracklet-phrase features $\{\mathcal{T}_i^{o},\mathcal{P}^o_i\}$ and $C_h$ by Eq.~\ref{eq:loss_obj};
        \STATE $\mathcal{P}^a$ $\leftarrow$ Get action phrase features with $\theta_M(a)$;
        \STATE $\mathcal{T}^a$ $\leftarrow$ Get action tracklet features with $\theta_A(\mathcal{Q},\{\mathcal{T}^o_i\}_{i=1}^N)$;
        \STATE $\mathcal{L}_{\text{act}}$ $\leftarrow$ Compute \textit{Relation-Guided Action Alignment} with action tracklet-phrase features $\{\mathcal{T}^{a},\mathcal{P}^a\}$ and $C_h$ by Eq.~\ref{eq:loss_act};
        \vspace{1mm}
        \STATE Update $\theta_M$ and $\theta_A$ by $\mathcal{L}_\text{total}$ in Eq.~\ref{eq:loss_total}
    \ENDFOR
\end{algorithmic}
\endgroup
\end{algorithm}
In Algorithm~\ref{algo}, we provide an overview of our training process for our proposed~\textit{SANTA}.~\textit{SANTA} employs hallucinative self-augmentation to generate the hallucinative captions serving as contrastive negatives during training. Subsequently, by jointly optimizing the proposed tracklet-phrase contrastive alignment $\mathcal{L}_{\text{obj}}+\mathcal{L}_{\text{act}}$ along with  $\mathcal{L}_{\text{video}}+\mathcal{L}_{\text{g}}$ from previous work~\cite{hacl} , \textit{SANTA} further achieves the enhanced object and action faithfulness. The overall training objective is as follows:
\begin{equation}\label{eq:loss_total}
\mathcal{L}_{\text{total}} =\alpha(\mathcal{L}_{\text{obj}}+\mathcal{L}_{\text{act}})+\beta\mathcal{L}_{\text{video}}+\mathcal{L}_{\text{g}}.
\end{equation}
Here, $\mathcal{L}_{\text{g}}$ is the cross-entropy loss used for the next token prediction to ensure the model retains its original capability of generating fluent outputs. $\alpha$ and $\beta$ are hyperparameters to control the relative contribution of the tracklet-phrase and video-caption contrastive alignments to $\mathcal{L}_{\text{total}}$.

\label{sec:3.4}
\section{Experiments}

\begingroup
\begin{table*}[ht]
\centering
\fontsize{8.5}{9.5}\selectfont
\setlength{\tabcolsep}{1.1mm}
\vspace{-4mm}

\caption{Evaluation on both object and action hallucinations with existing methods on MiraData-9k~\cite{mirabench} across three types of video captions: overall content, main object, and background. \textbf{Bold} and \underline{underline} indicate the best and second best results, respectively.}
\renewcommand{\arraystretch}{1.12}
\vspace{-3mm}

\begin{tabular}{l*{6}{c}*{6}{c}}
\toprule
\multirow{2}{*}{\textbf{Method}}
    & \multicolumn{6}{c}{\textbf{HalFscore~\cite{perturbollava}}}
    & \multicolumn{6}{c}{\textbf{weighted-HalFscore}} \\
\cmidrule(lr){2-7} \cmidrule(lr){8-13}
    & {\footnotesize \textbf{Hal$_\text{obj}$↓}}
    & {\footnotesize \textbf{Hal$_\text{act}$↓}}
    & {\footnotesize \textbf{Cov$_\text{obj}$↑}}
    & {\footnotesize \textbf{Cov$_\text{act}$↑}}
    & {\footnotesize \textbf{F1$_\text{obj}$↑}}
    & {\footnotesize \textbf{F1$_\text{act}$↑}}
    & {\footnotesize \textbf{Hal$_\text{obj}$↓}}
    & {\footnotesize \textbf{Hal$_\text{act}$↓}}
    & {\footnotesize \textbf{Cov$_\text{obj}$↑}}
    & {\footnotesize \textbf{Cov$_\text{act}$↑}}
    & {\footnotesize \textbf{F1$_\text{obj}$↑}}
    & {\footnotesize \textbf{F1$_\text{act}$↑}} \\
\midrule
\multicolumn{13}{c}{\textcolor{gray}{Overall Content Captioning}} \\
LLaVA-Video~\cite{llavavideo} & 79.4 & 76.5 & 52.1 & \textbf{33.0} & 29.5 & 27.5 & 49.0 & 58.4 & 54.6 & 32.9 & 52.7 & 36.7 \\
$+$SFT~\cite{llavavideo}      & 72.0 & 67.6 & 49.7 & 23.3 & 35.8 & 27.1 & 44.9 & 54.6 & 54.7 & 22.9 & 54.9 & 30.4 \\
$+$DeCo~\cite{deco}           & 74.2 & 73.4 & 46.6 & 27.3 & 33.2 & 26.9 & 48.4 & 56.6 & 48.1 & 27.6 & 49.8 & 33.8 \\
$+$HALVA~\cite{halva}         & 78.9 & 75.0 & 50.5 & 29.4 & 29.8 & 27.0 & 48.7 & 58.0 & 53.5 & 28.5 & 52.4 & 34.0 \\
$+$HACL~\cite{hacl}           & 70.0 & 65.2 & 53.7 & 29.9 & 38.4 & 32.2 & 42.7 & 51.6 & 60.8 & 28.6 & 59.0 & 35.9 \\
$+$\textbf{SANTA (Ours)}      & \textbf{69.6} & \textbf{64.5} & \textbf{57.2} & \underline{32.6} & \cellcolor{cyan!15}\textbf{39.7} & \cellcolor{cyan!15}\textbf{34.0} & \textbf{40.9} & \textbf{50.3} & \textbf{64.1} & \underline{31.5} & \cellcolor{cyan!15}\textbf{61.5} & \cellcolor{cyan!15}\textbf{38.6} \\
\midrule
\multicolumn{13}{c}{\textcolor{gray}{Main Object Captioning}} \\
LLaVA-Video~\cite{llavavideo} & 83.1 & 75.4 & 43.1 & 23.8 & 24.3 & 24.1 & 53.2 & 58.1 & 46.1 & 24.3 & 46.4 & 30.8 \\
$+$SFT~\cite{llavavideo}      & 79.5 & 68.1 & 49.7 & 23.3 & 29.0 & 26.9 & 44.2 & 55.4 & 54.8 & 21.4 & 55.2 & 28.9 \\
$+$DeCo~\cite{deco}           & 80.9 & 73.5 & 37.7 & 20.7 & 25.3 & 23.2 & 53.6 & 57.2 & 39.5 & 20.9 & 42.7 & 28.1 \\
$+$HALVA~\cite{halva}         & 84.9 & 77.2 & 46.7 & 24.0 & 22.8 & 23.4 & 55.0 & 60.1 & 49.9 & 23.5 & 47.4 & 29.5 \\
$+$HACL~\cite{hacl}           & 79.4 & 67.9 & 52.3 & 26.3 & 29.6 & 28.9 & 45.6 & 55.8 & 60.8 & 25.7 & 57.5 & 32.5 \\
$+$\textbf{SANTA (Ours)}      & \textbf{77.7} & \textbf{65.7} & \textbf{53.7} & \textbf{27.1} & \cellcolor{cyan!15}\textbf{31.6} & \cellcolor{cyan!15}\textbf{30.3} & \textbf{43.3} & \textbf{54.6} & \textbf{61.7} & \textbf{26.6} & \cellcolor{cyan!15}\textbf{59.1} & \cellcolor{cyan!15}\textbf{33.5} \\
\midrule
\multicolumn{13}{c}{\textcolor{gray}{Background Captioning}} \\
LLaVA-Video~\cite{llavavideo} & 78.7 & 85.4 & 42.3 & 19.4 & 38.9 & 22.3 & 65.1 & 73.0 & 44.0 & 19.0 & 38.9 & 22.3 \\
$+$SFT~\cite{llavavideo}      & 63.3 & 70.1 & 43.0 & 18.2 & 39.6 & 22.6 & 63.4 & 70.7 & 38.2 & 14.3 & 37.4 & 19.2 \\
$+$DeCo~\cite{deco}           & 77.1 & 86.2 & 40.0 & 16.2 & 29.0 & 14.9 & 65.2 & 73.4 & 41.3 & 15.9 & 37.8 & 19.9 \\
$+$HALVA~\cite{halva}         & 81.6 & 84.5 & \textbf{50.0} & \textbf{22.9} & 26.9 & 18.5 & 67.0 & 74.4 & \textbf{51.5} & \textbf{22.8} & 40.2 & 24.2 \\
$+$HACL~\cite{hacl}           & \textbf{60.3} & \textbf{65.7} & 43.5 & 18.8 & 41.5 & 24.3 & 60.4 & 68.5 & 44.7 & 18.4 & 42.0 & 23.2 \\
$+$\textbf{SANTA (Ours)}      & \underline{62.2} & \underline{68.3} & \underline{47.2} & \underline{22.8} & \cellcolor{cyan!15}\textbf{42.3} & \cellcolor{cyan!15}\textbf{25.8} & \textbf{60.2} & \textbf{68.4} & \underline{48.6} & \underline{22.3} & \cellcolor{cyan!15}\textbf{43.8} & \cellcolor{cyan!15}\textbf{26.1} \\
\bottomrule
\end{tabular}
\vspace{-1.5mm}
\label{tab:main}
\end{table*}
\endgroup

\begin{table*}[!ht]
\centering
\begin{minipage}[t]{0.49\textwidth}
\captionsetup{type=table} 
\caption{Quantitative comparisons with hallucination mitigation methods on video captioning using FactVC~\cite{factvc}.}
\label{tab:factvc}
\vspace{-3mm}

\makebox[\linewidth]{
    \scalebox{0.85}{
        \begin{tabular}{l ccc}
        \toprule
        \textbf{Method} & \textbf{$\text{FactVC}_{\text{prec.}}$}$\uparrow$ & \textbf{$\text{FactVC}_{\text{rec.}}$}$\uparrow$ & \textbf{$\text{FactVC}_{\text{F1}}$}$\uparrow$\\
        \midrule
        LLaVA-Video~\cite{llavavideo} & 36.4 & 39.9 & 37.9 \\
        $+$SFT~\cite{llavavideo} & 38.2 & 41.7 & 39.7 \\
        $+$DeCo~\cite{deco} & 36.1 & 39.6 & 37.6 \\
        $+$HALVA~\cite{halva} & 38.2 & 41.6 & 39.7 \\
        $+$HACL~\cite{hacl} & 38.9 & 42.4 & 40.5 \\
        $+$\textbf{SANTA (Ours)} & \textbf{39.4} & \textbf{42.9} & \cellcolor{cyan!15}\textbf{40.9} \\
        \bottomrule
        \end{tabular}
        }
}
\vspace{-3.5mm}

\end{minipage}
\hfill
\begin{minipage}[t]{0.49\linewidth}
\captionsetup{type=table} 
\caption{Quantitative evaluation of both object and action hallucinations on video question answering using VidHal~\cite{vidhal}.}
\label{tab:vidhal}
\vspace{-3mm}

\makebox[\linewidth]{
    \scalebox{0.85}{
        \begin{tabular}{l ccc}
        \toprule
        \textbf{Method} & \textbf{Acc$_\text{obj}$}$\uparrow$ & \textbf{Acc$_\text{act}$}$\uparrow$ & \textbf{Avg. Acc}$\uparrow$ \\
        \midrule
        LLaVA-Video~\cite{llavavideo}     & 84.3 & 83.1  & 83.7 \\
        $+$SFT~\cite{llavavideo} & 36.3 & 65.6  & 51.0 \\
        $+$DeCo~\cite{deco}      & 82.8 & 84.7& 83.8\\
        $+$HALVA~\cite{halva}    & 79.9 & 80.9  & 80.4    \\
        $+$HACL~\cite{hacl}      & 79.9 & 84.2  & 82.1    \\
        $+$\textbf{SANTA (ours)} & \textbf{86.3} & \textbf{85.8} & \cellcolor{cyan!15}\textbf{86.1} \\
        \bottomrule
        \end{tabular}
    }
}
\vspace{-3.5mm}

\end{minipage}
\end{table*}
\begin{figure*}[h]
    \centering
    
    \vspace{-7mm}
    \includegraphics[width=0.95\linewidth, height=0.45\linewidth, keepaspectratio=false]{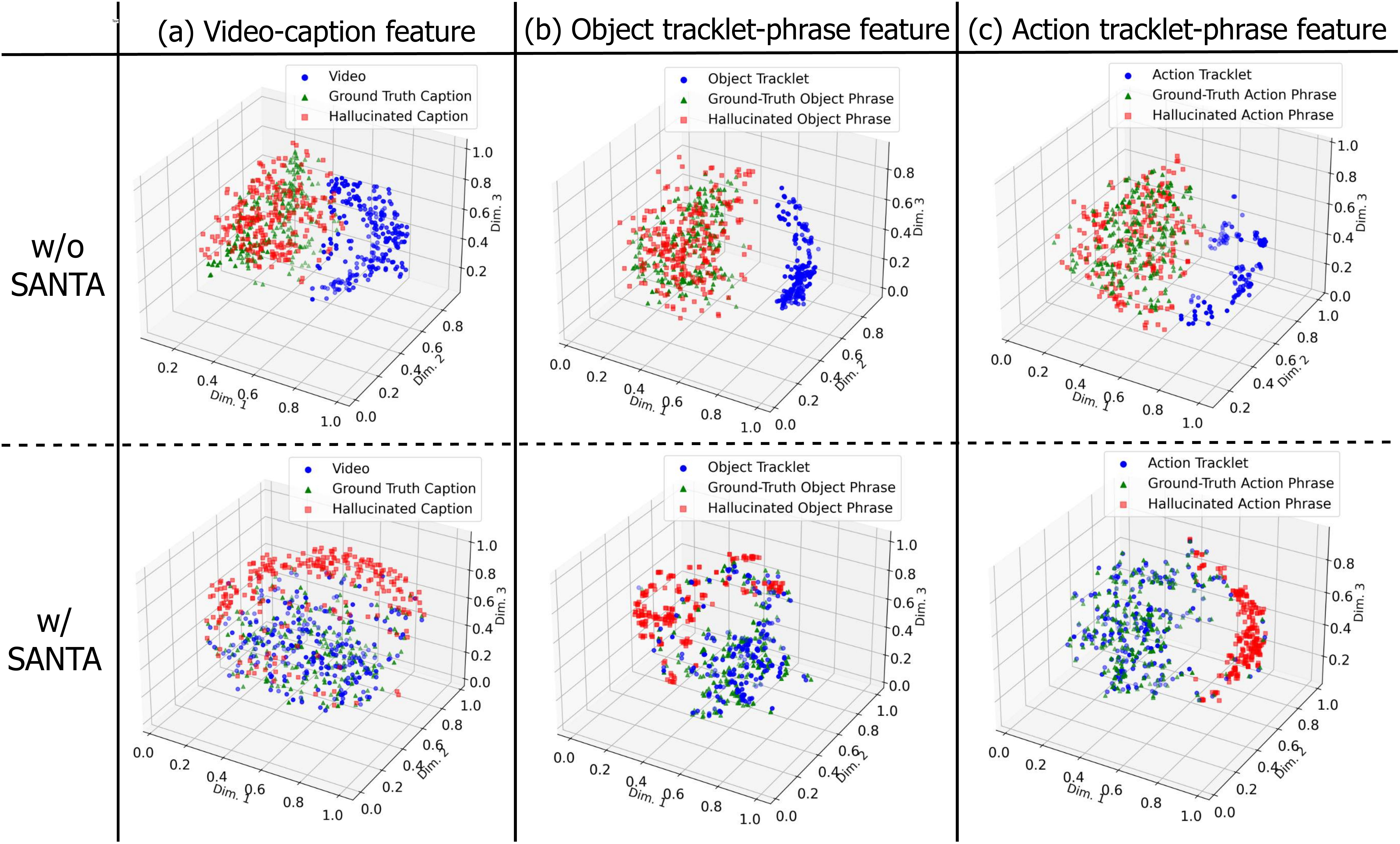}
    \vspace{-3mm}
    \caption{t-SNE visualization of the latent features of (a) video and caption, (b) object tracklets and phrases, and (c) action tracklets and phrases. For the w/o ~\textit{SANTA} setting, we directly visualize features from LLaVA-Video~\cite{llavavideo}. Upon training with \textit{SANTA} LLaVA-Video~\cite{llavavideo} improves the alignment between visual-language modalities while exempting from the hallucinative captions.}
    \vspace{-2mm}
    \label{fig:tsne}
\end{figure*}

\begin{figure*}[t]
    \centering
    \includegraphics[width=0.98\linewidth, height=0.32\linewidth]{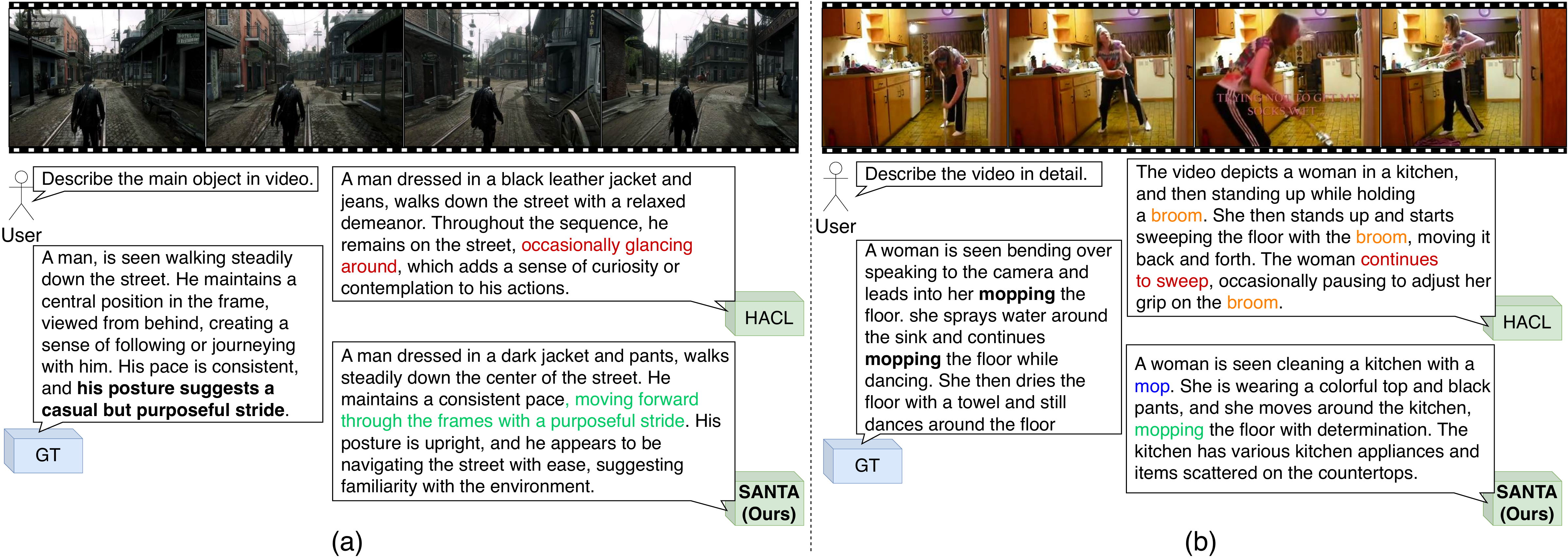}
    \vspace{-3mm}
    \caption{Qualitative comparison of video captions predicted by HACL~\cite{hacl} and \textit{SANTA}. Note that words highlighted in \textcolor{ForestGreen}{green} indicate action faithfulness, while those in \textcolor[HTML]{C23B22}{red} indicate action hallucination. Similarly, words in \textcolor{blue}{blue} represent object faithfulness, whereas those in \textcolor{orange}{orange} denote object hallucination. The examples of (a) and (b) are sampled from MiraData-9k~\cite{mirabench} and FactVC~\cite{factvc}, respectively.}   
    \label{fig:qualitative_result}
    \vspace{-4.5mm}
\end{figure*}

\subsection{Experimental Setup}
\paragraph{Training Data.} 
We use MiraData~\cite{mirabench} as our training data, which contains high-quality and detailed captions for each video. Our training data consists of 42715 videos with an average of 72.1 seconds per video. Each video is annotated with three types of captions: \textit{overall captions}, \textit{main object captions}, and \textit{background captions}. The average caption length across all types is 318 words. More details are in the supplementary material.

\vspace{-2mm}
\paragraph{Evaluation Benchmarks.}
To assess both the object and action hallucination in the textual description generated by multimodal large language models (MLLMs), we evaluate performance on video captioning datasets (i.e., MiraData-9k~\cite{mirabench} and FactVC~\cite{factvc}) and multiple-choice video question answering datasets (i.e., VidHal~\cite{vidhal}). Note that FactVC is composed of curated samples from ActivityNet~\cite{actnet} and YouCook2~\cite{youcook2}, and that the MiraData-9k and the MiraData used during training do not overlap. Furthermore, we assess video description capability on Dream1k~\cite{dream1k} following hallucination mitigation.

\vspace{-3mm}
\paragraph{Hallucination Evaluation Metrics.}
\label{exp:hallu_metrics}

To evaluate video hallucinations, existing metrics such as FactVC~\cite{factvc} and HalFscore~\cite{perturbollava} are adopted. \textbf{FactVC$_{\text{F1}}$}~\cite{factvc} utilizes CLIP~\cite{clip} embeddings to computes an F1 score to assess hallucination in generated captions, while \textbf{HalFscore}~\cite{perturbollava} computes an F1 score based on exact string matching. The latter comprises two sub-scores: a hallucination-score (precision, denoted as $\text{Hal}$) and a coverage-score (recall, denoted as $\text{Cov}$), reflecting the extent of hallucination and the comprehensiveness of detail coverage, respectively, with subscripts ``Obj'' and ``Act'' denoting object- and action-level evaluations (e.g., F1$_\text{obj}$ and F1$_\text{act}$). However, HalFscore~\cite{perturbollava} fails to recognize synonyms and potentially misjudges hallucinations, while FactVC$_{\text{F1}}$~\cite{factvc} cannot be directly applied to MiraData-9k~\cite{mirabench}, due to the token length limitation of CLIP~\cite{clip} (i.e., 78 tokens). 


To address these limitations, we extend HalFscore ~\cite{perturbollava} by modifying its exact string matching to cosine-similarity for synonym awareness in Hal$_{\text{obj}}$ and Hal$_{\text{act}}$ while by incorporating tf-idf~\cite{tfidf} weighting to better assess the importance of tokens in Cov$_{\text{obj}}$ and Cov$_{\text{act}}$. We term this improved metric as \textbf{weighted-HalFscore}, comprising two sub-scores, formally defined as follows:


\vspace{-2mm}
\begin{equation}
\text{Hal}=1-\frac{1}{|\mathcal{P}|} \sum_{t_i \in \mathcal{P}} \max_{t_j \in \mathcal{G}} \left\{ \text{cos}(t_i, t_j) \right\},    
\end{equation}
\begin{equation}
\text{Cov}=\frac{\sum_{t_i \in (\mathcal{P} \cap \mathcal{G})} \text{tf-idf}(t_i)}{\sum_{t_j \in \mathcal{G}} \text{tf-idf}(t_j)}
\end{equation}
where $\mathcal{P}$ and $\mathcal{G}$ are the sets of predicted and ground-truth object (or action) tokens, and $\text{tf-idf}(t_i)$ is the tf-idf weight for token $t_i$. The overall F1 score is then computed as:

\vspace{-3mm}

\begin{equation}
\text{F1}=\frac{2 \times (1-\text{Hal})\times\text{Cov}}{(1-\text{Hal})+\text{Cov}}.    
\end{equation}

\vspace{-4mm}



\paragraph{Implementation Details.}
We adopt the LLaVA-Video$_\text{7B}$~\cite{llavavideo} and Qwen2.5-VL$_\text{7B}$~\cite{qwen2_5vl}, as our base models. Training is conducted with a learning rate of 6e-5, a batch size of 64, and 64 frames uniformly sampled from each video. All frames are resized to 384 pixels, and the models are trained for 2,000 steps. The action-squeezer is initialized from the perceiver model~\cite{paxion} pre-trained on video dynamics modeling. This module consists of a single cross-attention layer followed by a two-layer feedforward network, employing $N_{Q}=16$ learnable queries, adopted from the configuration of~\cite{paxion}. Hyperparameters $\alpha$ and $\beta$ are set to 0.25 and 0.5, respectively.

\subsection{Quantitative Evaluation}
\paragraph{\mbox{Video Hallucination Evaluation with Captioning.}}
In Tab.~\ref{tab:main} and Tab.~\ref{tab:factvc}, we evaluate \textit{SANTA} on MiraData-9k~\cite{mirabench} and FactVC~\cite{factvc} against prior SOTA methods. As shown in Tab.~\ref{tab:main}, \textit{SANTA} achieves the highest F1$_{\text{obj}}$ and F1$_{\text{act}}$ of Falscore~\cite{perturbollava} across three captioning subtasks, with average gains of +4.02\%/+5.54\% over the second-best method, and +3.77\%/+7.7\% for weighted-Falscore. Tab.~\ref{tab:factvc} further confirms these improvements, w/ \textit{SANTA} reaching the highest precision (FactVC$_\text{prec.}$), recall (FactVC$_\text{rec.}$), and F1 (FactVC$_\text{F1}$). These consistent results highlight that \textit{SANTA} effectively minimize hallucinations while maximizing coverage of relevant tokens from the provided videos.


\begingroup
\begin{table}[ht]
\centering
\fontsize{8.5}{9.5}\selectfont
\setlength{\tabcolsep}{2mm}
\vspace{-6mm}
\caption{Ablation of~\textit{SANTA} on different MLLM backbones. We report the averaged F1 score of MiraData-9k~\cite{mirabench} (HalFscore~\cite{perturbollava} and Weighted-HalFscore) to evaluate the effectiveness in mitigating both object and action hallucinations.}
\renewcommand{\arraystretch}{1.12}
\vspace{-3mm}

\begin{tabular}{lcccc}
\toprule
\multirow{2}{*}{\textbf{Method}}
    & \multicolumn{2}{c}{\textbf{HalFscore~\cite{perturbollava}}}
    & \multicolumn{2}{c}{\textbf{weighted-HalFscore}} \\
\cmidrule(lr){2-3} \cmidrule(lr){4-5}
    & {\footnotesize \textbf{\shortstack{Avg. \\ F1$_\text{Obj}$↑}}}
    & {\footnotesize \textbf{\shortstack{Avg. \\ F1$_\text{Act}$↑}}}
    & {\footnotesize \textbf{\shortstack{Avg. \\ F1$_\text{Obj}$↑}}}
    & {\footnotesize \textbf{\shortstack{Avg. \\ F1$_\text{Act}$↑}}} \\
\midrule
\multicolumn{5}{c}{\textcolor{gray}{Overall Content Captioning}} \\
LLaVA-Video~\cite{llava}                 & 29.5 & 27.5 & 52.7 & 36.7 \\
$+$\textbf{SANTA (Ours)}                 & \cellcolor{cyan!15}\textbf{39.7} & \cellcolor{cyan!15}\textbf{34.0} & \cellcolor{cyan!15}\textbf{61.5} & \cellcolor{cyan!15}\textbf{38.6} \\
\arrayrulecolor{black!20}\midrule\arrayrulecolor{black}
Qwen2.5-VL~\cite{qwen2_5vl}              & \textbf{44.3} & 34.7 & 60.9 & 40.7 \\
$+$\textbf{SANTA (Ours)}                 & \cellcolor{cyan!15}42.1 & \cellcolor{cyan!15}\textbf{36.7} & \cellcolor{cyan!15}\textbf{65.5} & \cellcolor{cyan!15}\textbf{41.1} \\
\midrule
\multicolumn{5}{c}{\textcolor{gray}{Main Object Captioning}} \\
LLaVA-Video~\cite{llava}                 & 24.3 & 24.1 & 46.4 & 30.8 \\
$+$\textbf{SANTA (Ours)}                 & \cellcolor{cyan!15}\textbf{31.6} & \cellcolor{cyan!15}\textbf{30.3} & \cellcolor{cyan!15}\textbf{59.1} & \cellcolor{cyan!15}\textbf{33.5} \\
\arrayrulecolor{black!20}\midrule\arrayrulecolor{black}
Qwen2.5-VL~\cite{qwen2_5vl}              & \textbf{34.4} & 29.7 & 55.3 & 36.8 \\
$+$\textbf{SANTA (Ours)}                 & \cellcolor{cyan!15}32.8 & \cellcolor{cyan!15}\textbf{32.3} & \cellcolor{cyan!15}\textbf{63.4} & \cellcolor{cyan!15}\textbf{37.0} \\
\midrule
\multicolumn{5}{c}{\textcolor{gray}{Background Captioning}} \\
LLaVA-Video~\cite{llava}                 & 38.9 & 22.3 & 38.9 & 22.3 \\
$+$\textbf{SANTA (Ours)}                 & \cellcolor{cyan!15}\textbf{42.3} & \cellcolor{cyan!15}\textbf{25.8} & \cellcolor{cyan!15}\textbf{43.8} & \cellcolor{cyan!15}\textbf{26.1} \\
\arrayrulecolor{black!20}\midrule\arrayrulecolor{black}
Qwen2.5-VL~\cite{qwen2_5vl}              & 33.4 & 23.3 & 45.4 & 28.4 \\
$+$\textbf{SANTA (Ours)}                 & \cellcolor{cyan!15}\textbf{41.8} & \cellcolor{cyan!15}\textbf{27.4} & \cellcolor{cyan!15}\textbf{45.9} & \cellcolor{cyan!15}\textbf{29.2} \\
\bottomrule
\end{tabular}
\label{tab:miradata_diff_backbone}
\end{table}
\vspace{-5mm}
\endgroup

\begin{table}[t]
\centering
\vspace{-2mm}
\caption{Ablation of~\textit{SANTA} on key components. All metrics are used as described in Tab.~\ref{tab:miradata_diff_backbone}.}
\vspace{-3mm}

\resizebox{\columnwidth}{!}{
\begin{tabular}{c c c c c c c c}
\toprule
\multirow{3}{*}{\footnotesize{$\mathcal{L}_\text{g}$$+$$\mathcal{L}_\text{video}$}}
  & \multirow{3}{*}{$\mathcal{L}_\text{obj}$}
  & \multirow{3}{*}{$\mathcal{L}_\text{act}$}
  & \multirow{3}{*}{$\mathcal{C}_h$}
  & \multicolumn{2}{c}{\textbf{HalFscore~\cite{perturbollava}}}
  & \multicolumn{2}{c}{\textbf{weighted-HalFscore}} \\
\cmidrule(lr){5-6} \cmidrule(lr){7-8}
  & & & 
  & \textbf{\shortstack{Avg. \\ F1$_\text{Obj}$↑}} & \textbf{\shortstack{Avg. \\ F1$_\text{Act}$↑}}
  & \textbf{\shortstack{Avg. \\ F1$_\text{Obj}$↑}} & \textbf{\shortstack{Avg. \\ F1$_\text{Act}$↑}} \\
\midrule
\checkmark & --         & --         & --          & 34.8 & 25.5 & 51.1 & 29.3 \\
\checkmark & \checkmark & --         & --          & 36.7 & 28.9 & 53.3 & 31.1 \\
\checkmark & \checkmark & \checkmark & --          & 37.0 & 29.9 & 53.9 & 32.3 \\
\checkmark & \checkmark & \checkmark & \checkmark 
           & \cellcolor{cyan!15}\textbf{37.9} 
           & \cellcolor{cyan!15}\textbf{30.0} 
           & \cellcolor{cyan!15}\textbf{54.8} 
           & \cellcolor{cyan!15}\textbf{32.7} \\
\bottomrule
\end{tabular}
}
\vspace{-3mm}
\label{tab:ablation_santa}
\end{table}

\paragraph{\mbox{Video Hallucination Evaluation with QA.}}
In Tab.~\ref{tab:vidhal}, \textit{SANTA} achieves the highest accuracy on both object- and action-level hallucination QA tasks, surpassing all baselines. This shows that beyond caption hallucination mitigation (Tab.~\ref{tab:main} and \ref{tab:factvc}), \textit{SANTA} also provides the most accurate answers for hallucination-focused video QA.


\subsection{tSNE Visualization}
In Fig.~\ref{fig:tsne}, we visualize the output visual and language features by the LLaVA-Video~\cite{llavavideo} under two conditions: with and without \textit{SANTA}. Each tSNE subfigure uses 250 MiraData-9k~\cite{mirabench} pairs: (a) video-caption, (b) object tracklet-phrase, and (c) action tracklet-phrase. Hallucinated captions are generated via hallucinative self-augmentation (introduced in Eq.~\ref{eq:bootstrapped_hn}) before \textit{SANTA} training. Above the dotted line, a clear vision-language gap and substantial overlap between ground-truth and hallucinated language features are observed. After applying \textit{SANTA} (below the line), the gap narrows and hallucinated language features become more clearly distinguishable from the ground-truth.


\subsection{Qualitative Evaluation}
In Fig.\ref{fig:qualitative_result}, we compare captions from \textit{SANTA}, HACL\cite{hacl}, and the ground truth on MiraData-9k~\cite{mirabench} (a) and FactVC~\cite{factvc} (b). In (a), \textit{SANTA} generates an action-faithful caption closely matching the ground truth, while HACL~\cite{hacl} hallucinates the action “occasionally glancing around” (red), which is absent in the video. In (b), \textit{SANTA} accurately captures both object and action (“mop,” “mopping”), whereas HACL hallucinates both an object (“broom,” orange) and an action (“continues to sweep,” red).


\subsection{Analysis}
\paragraph{\mbox{Ablation on Different Backbones.}}
Beyond the strong performance of \textit{SANTA} on LLaVA-Video~\cite{llavavideo} (in Tab.~\ref{tab:main}, \ref{tab:factvc}, and \ref{tab:vidhal}), Tab.~\ref{tab:miradata_diff_backbone} shows that adding \textit{SANTA} to LLaVA-Video consistently boosts both object- and action-level scores across all three MiraData-9k~\cite{mirabench} captioning subtasks, demonstrating its effectiveness even on a strong baseline. Moreover, applying \textit{SANTA} to Qwen2.5-VL~\cite{qwen2_5vl} yields further notable gains, particularly in action-level and weighted-HalFscore metrics, confirming the robustness and general applicability of our method across different MLLM architectures.


\vspace{-3mm}
\paragraph{\mbox{Ablation on Key Components.}}
In Tab.~\ref{tab:ablation_santa}, introducing region‐level object alignment into LLaVA-Video~\cite{llavavideo} boosts F1$_{\text{obj}}$/F1$_{\text{act}}$ of HalFscore~\cite{perturbollava} by +5.5\%/+13.3\% (weighted-HalFscore +4.3\%/+6.1\%) over $\mathcal{L}_\text{g}+\mathcal{L}_{\text{video}}$. Adding action alignment bring further gains of +0.8\%/+3.5\% (+1.1\%/+3.9\%), while the complete \textit{SANTA} achieves the best results, adding +2.4\%/+0.3\% (+1.7\%/+1.2\%).


\vspace{-3mm}
\paragraph{\mbox{Video Description Evaluation.}}
To demonstrate the benefit of enhancing object and action faithfulness, we apply \textit{SANTA} to LLaVA-Video~\cite{llavavideo} and observe improved video description performance on Dream1k~\cite{dream1k}. The overall F1-score improves from 32.5 to 32.7, confirming that increasing faithfulness reduces hallucinations and improves descriptive ability. Detailed results for each video type are in the supplementary material.

\begin{figure}[t]
  \centering
  \vspace{-10mm}
  \includegraphics[width=1.0\linewidth, height=1.28\linewidth, keepaspectratio=false]{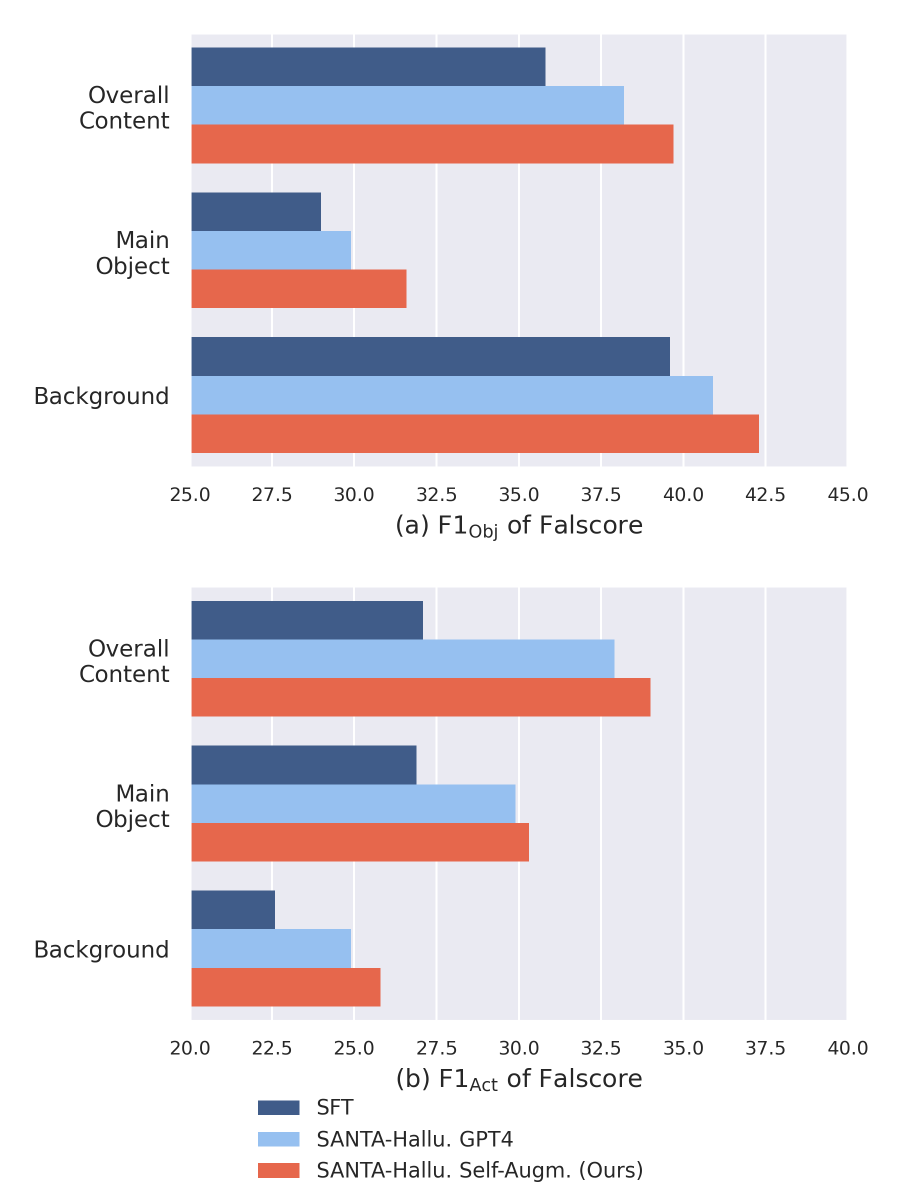}
  \vspace{-6mm}
  \caption{Ablation study of hallucinative self-augmentation scheme. We ablate this scheme by replacing it with negatives generated directly by a text-only LLM (i.e., GPT-4~\cite{gpt4}), following the prompting setup of HACL~\cite{hacl}. We report $\text{F1}_{\text{Obj}}$ and $\text{F1}_{\text{Act}}$ of HalFscore~\cite{perturbollava} to evaluate the effectiveness of mitigating object and action hallucinations, respectively.
  }
  \vspace{-4mm}
  \label{fig:ablation_hn}
\end{figure}

\vspace{-3mm}
\paragraph{\mbox{Impact of Hallucinative Self-Augmentation Scheme.}}
In Fig.~\ref{fig:ablation_hn}, we present F1$_{\text{obj}}$ (a) and F1$_{\text{act}}$ (b) for hallucination evaluations across overall content, main object, and background captioning on Miradata-9k~\cite{mirabench}. Our hallucinative self-augmentation scheme generates negative samples that better reflect the hallucinations that the MLLM itself is prone to producing. As a result, our approach not only surpasses the simple supervised fine-tuning baseline (SFT), but also outperforms the variant where \textit{SANTA} uses hard negatives generated by GPT4o~\cite{gpt4o} (denoted as SANTA-Hallu. GPT4), which adopts exactly the same negatives generation strategy used in HACL~\cite{hacl} and HALVA~\cite{halva}. This demonstrates that leveraging self-generated hallucinative samples enables more effective hallucination mitigation than relying solely on GPT4o-generated negatives.


\section{Conclusion}
In this paper, we propose a self-augmented contrastive alignment (\textit{SANTA}) framework designed to mitigate hallucinations in multimodal LLMs for video understanding tasks. \textit{SANTA} employs a hallucinative self-augmentation scheme to expose and counteract spurious correlations in generated descriptions while encouraging the focus on visual facts through tracklet-phrase contrastive alignment. By explicitly linking regional objects and relation-guided actions with their corresponding visual and temporal phrases, \textit{SANTA} enhances both object and action faithfulness in the output captions. Extensive experiments demonstrate that \textit{SANTA} significantly reduces hallucinations compared to existing methods, achieving state-of-the-art performance on hallucination examination benchmarks.
\section{Acknowledgments}
This work is supported in part by the National Science and Technology Council via grant NSTC 114-2221-E-002-056-MY2 and NSTC 114-2640-E-002-006. We also thank the National Center for High-performance Computing (NCHC) for providing computational and storage resources.
{
    \small
    \bibliographystyle{ieeenat_fullname}
    \bibliography{main}
}

\clearpage
\setcounter{page}{1}




\begin{table*}[t]
    \centering
    \caption{Quantitative comparison of hallucination mitigation methods on the Dream1k~\cite{dream1k} benchmark. \textbf{Bold} and \underline{underline}, respectively, indicate the best and second best results achieved by these methods over the LLaVA-Video~\cite{llavavideo}.}
    \label{tab:dream1k}
    \resizebox{0.75\textwidth}{!}{%
    \begin{tabular}{l c c c c c c}
        \toprule
        \textbf{Method}
            & \textbf{Animation}
            & \textbf{Live action}
            & \textbf{Short}
            & \textbf{Stock}
            & \textbf{Youtube}
            & \textbf{Overall} \\
        \midrule
        LLaVA-Video~\cite{llavavideo}  & \textcolor{gray}{27.6} & \textcolor{gray}{31.4} & \textcolor{gray}{33.4} & \textcolor{gray}{36.7} & \textcolor{gray}{33.0} & \textcolor{gray}{32.5} \\
        \cmidrule(lr){2-7} 
        +SFT~\cite{llavavideo}         & 19.9 & 27.7 & 28.3 & 36.0 & 33.0 & 29.2 \\
        +HALVA~\cite{halva}            & 26.2 & \textbf{32.3} & \textbf{34.6} & 38.2 & 31.8 & 32.6 \\
        +HACL~\cite{hacl}              & 23.9 & 28.1 & 30.0 & 37.5 & 33.2 & 30.7 \\
        +SANTA (Ours)                  & \textbf{24.7} & \underline{31.0} & \underline{31.8} & \textbf{41.2} & \textbf{33.4} & \textbf{32.7} \\
        \bottomrule
    \end{tabular}
    }
\end{table*}

\section{Additional Experiment Setups}

\subsection{Training Data}
~\label{suppl:train_data}
Here, we provide a more detailed explanation of the three captioning subtasks in MiraData-9k~\cite{mirabench}. The first is overall captioning, which provides a high-level summary of the video's main content. The second is main object captioning, which focuses on describing the primary objects present in the video. The last is background captioning, which provides contextual information about the surrounding environment.


~\label{suppl:train_setups}


\label{suppl:hallu_self_augm}
We describe the process of obtaining the ground-truth action and object token set using GPT-4o~\cite{gpt4o} as parser, following the prompting templates presented in Fig.~\ref{fig:prompt_template}. As mentioned in~\ref{sec:3.2}, we parse the ground-truth captions to extract structured relation triples. Specifically, we employ GPT-4o~\cite{gpt4o} as a parser to derive each triple, consisting of an action verb and two associated object nouns, which we refer to as "related object instances" corresponding to the action verb. Thus, in our setup, $N$ is set to 2. After extracting these ground-truth token sets, we use WordNet~\cite{wordnet} to expand them by incorporating both synonyms (e.g., kid'' for child'' and get up'' for stand'') and hypernyms (e.g., person'' for child'' and move'' for stand''). Since each video is described using three enriched captions, the extracted tokens comprehensively represent the visual facts depicted in the video. As a result, this augmentation process allows self-augmented hallucinative captions to capture the hallucination tendencies of the generated MLLM while ensuring that no visual information outside the video's content is introduced.

\paragraph{\mbox{Tracklet-Phase Contrastive Alignment.}}
As described in~\ref{sec:3.3},  to obtain the object tracklets, we employ Grounded-SAM2~\cite{groundingdino1.5, sam2} to generate segmentation masks for each object instance, based on object information available in ground-truth captions. Using these masks, we crop the corresponding object regions, allowing us to identify which visual tokens contain object-specific information. These identified visual tokens are then used for region-level object alignment and relation-guided action alignment. The entire process is conducted offline.

\subsection{Additional Evaluation Setups}
In this subsection, we describe how we use MiraData-9k~\cite{mirabench}, a video-caption benchmark, to evaluate object and action hallucinations. As mentioned in Sec.\ref{suppl:hallu_self_augm}, we parse the ground-truth captions to extract sets of ground-truth object and action tokens. Using these sets, we then apply the hallucination evaluation metrics (i.e., Hal$_{\text{Act}}$, Hal$_{\text{Obj}}$, Cov$_{\text{Act}}$, Cov$_{\text{Obj}}$, F1$_{\text{Act}}$, and F1$_{\text{Obj}}$ for both HalScore~\cite{perturbollava} and weighted-HalScore) introduced in Sec.\ref{exp:hallu_metrics} to assess hallucination mitigation methods.

\subsection{Comparisions} 
~\label{suppl:comparision}
We compare \textit{SANTA} to a simple baseline supervised fine-tuning baseline, and several the state-of-the-art visual hallucination mitigation methods, including DeCo~\cite{deco}, HALVA~\cite{halva}, and HACL~\cite{hacl}. For all methods, we utilize their publicly available codebases and adhere to their reported training configurations. To ensure a fair comparison, we align their training dataset, training and inference hyperparameter settings for all the methods, matching those used by LLaVA-Video~\cite{llavavideo} and Qwen2.5-VL~\cite{qwen2_5vl}.

\section{Additional Quantitative Results}
~\label{suppl:quant}

\paragraph{More Detailed Results on Video Description.}
Tab.~\ref{tab:dream1k} presents a quantitative comparison of various hallucination mitigation methods on Dream1k~\cite{dream1k}, a challenging video description benchmark comprising videos from five diverse sources: animation movies, live-action movies, TikTok-style videos, stock videos, and YouTube videos. The evaluation metric uses F1-score, calculated by the LLM-based evaluator, AutoDQ~\cite{dream1k}. As shown in Tab.~\ref{tab:dream1k}, \textit{SANTA} achieves the highest overall F1-score, outperforming existing methods such as HALVA~\cite{halva} and HACL~\cite{hacl}. This demonstrates that \textit{SANTA} not only preserves but also enhances the video captioning capabilities of MLLMs. Notably, \textit{SANTA} surpasses the original LLaVA-Video~\cite{llavavideo} baseline, further validating its effectiveness.

\paragraph{Results on General Video Question Answering.}
In Tab.~\ref{tab:videomme}, we present a quantitative comparison of different hallucination mitigation methods on the VideoMME~\cite{videomme} benchmark, evaluated under both with and without subtitles settings. VideoMME is a multiple-choice QA benchmark that spans a diverse range of video lengths, from short clips (11 seconds) to long-form content (up to 1 hour). The evaluation metric is accuracy, where higher scores indicate better video understanding performance. 

As presented in Tab.~\ref{tab:videomme}, when applied to the same LLaVA-Video backbone, \textit{SANTA} achieves higher QA accuracy than HACL~\cite{hacl}, improving the overall performance by +2.7 (with subtitles) and +0.9 (without subtitles). However, both methods still exhibit a slight decrease compared to the original backbone, which is a shared limitation of current hallucination mitigation approaches. We hypothesize that this stems from using only video captioning data (e.g., MiraData-9k~\cite{mirabench}) during post-training, without leveraging any video QA data. Thus, Appropriately incorporating a portion of this QA data could preserve the backbone’s QA capability and, when combined with the hallucination reduction achieved by \textit{SANTA}, further enhance overall performance beyond the original backbone.


\begin{table*}[t]
    \centering
    \caption{uantitative comparison of hallucination mitigation methods on the VideoMME~\cite{videomme} benchmark, with and without subtitles. \textbf{Bold} and \textbf{underline}, respectively, indicate the best and second best results achieved by these methods over the LLaVA-Video~\cite{llavavideo}.}
    \label{tab:videomme}
    \resizebox{0.8\textwidth}{!}{%
    \begin{tabular}{l *{4}{cc}}
        \toprule
        \multirow{2}{*}{\textbf{Method}} 
            & \multicolumn{2}{c}{\textbf{Short}} 
            & \multicolumn{2}{c}{\textbf{Medium}} 
            & \multicolumn{2}{c}{\textbf{Long}} 
            & \multicolumn{2}{c}{\textbf{Overall}} \\
        \cmidrule(lr){2-3} \cmidrule(lr){4-5} \cmidrule(lr){6-7} \cmidrule(lr){8-9}
            & \textbf{with} & \textbf{without} 
            & \textbf{with} & \textbf{without} 
            & \textbf{with} & \textbf{without} 
            & \textbf{with} & \textbf{without} \\
        \midrule
        LLaVA-Video~\cite{llavavideo} 
            & \textcolor{gray}{80.1} & \textcolor{gray}{77.0}
            & \textcolor{gray}{66.6} & \textcolor{gray}{61.8}
            & \textcolor{gray}{58.3} & \textcolor{gray}{52.1}
            & \textcolor{gray}{68.3} & \textcolor{gray}{63.6} \\
        \noalign{\vskip 0pt}
        \cmidrule[0.1pt](l){2-9}
        \noalign{\vskip 0pt}
        +SFT~\cite{llavavideo} 
            & 54.8 & 56.8 
            & 45.4 & 45.9 
            & 35.3 & 35.0 
            & 45.2 & 45.9 \\
        +HACL~\cite{hacl} 
            & 73.1 & 69.9 
            & 62.3 & 57.9 
            & 53.2 & 51.4 
            & 62.9 & 59.7 \\
        +SANTA (Ours) 
            & \textbf{76.8} & \textbf{72.3} 
            & \textbf{64.6} & \textbf{59.8} 
            & \textbf{55.4} & \textbf{49.6} 
            & \textbf{65.6} & \textbf{60.6} \\
        \bottomrule
    \end{tabular}
    }
\end{table*}

\begin{table}[!t]
\centering
\fontsize{8.5}{9.5}\selectfont
\setlength{\tabcolsep}{2mm}
\renewcommand{\arraystretch}{1.12}
\caption{Ablation of~\textit{SANTA} w/ the quality of object tracklets. We report the averaged F1 score of MiraData-9k~\cite{mirabench} (HalFscore~\cite{perturbollava} and Weighted-HalFscore) to evaluate the effectiveness in mitigating both object and action hallucinations.}
\label{tab:ablate_vg}
\resizebox{\linewidth}{!}{%
\begin{tabular}{lcccc}
\toprule
\multirow{2}{*}{\textbf{Method}}
    & \multicolumn{2}{c}{\textbf{HalFscore~\cite{perturbollava}}}
    & \multicolumn{2}{c}{\textbf{weighted-HalFscore}} \\
\cmidrule(lr){2-3} \cmidrule(lr){4-5}
    & {\footnotesize \textbf{\shortstack{Avg. \\ F1$_\text{Obj}$↑}}}
    & {\footnotesize \textbf{\shortstack{Avg. \\ F1$_\text{Act}$↑}}}
    & {\footnotesize \textbf{\shortstack{Avg. \\ F1$_\text{Obj}$↑}}}
    & {\footnotesize \textbf{\shortstack{Avg. \\ F1$_\text{Act}$↑}}} \\
\midrule
\makecell[l]{HACL~\cite{hacl}} & 36.5 & 28.5 & 52.8 & 30.5 \\
\arrayrulecolor{black!20}\midrule\arrayrulecolor{black}
\makecell[l]{SANTA w/ G-SAM2 \\ ($t$=0.15)} & 37.2 & 29.4 & 54.0 & 31.6 \\
\arrayrulecolor{black!20}\midrule\arrayrulecolor{black}
\makecell[l]{SANTA w/ G-SAM2 \\ ($t$=0.25)} & \textbf{37.9} & \textbf{30.0} & \textbf{54.8} & \textbf{32.7} \\
\bottomrule
\end{tabular}
}
\end{table}

\paragraph{Analysis of Robustness with the Impact of Object Tracklet Quality.}
We use the SOTA video tracker Grounded-SAM2~\cite{groundingdino1.5, sam2} to extract object tracklets. To analyze the robustness against possible tracking errors, we now ablate \textit{SANTA} by training with noisy object tracklets, by lowering the detection confidence threshold (from default 0.25 to 0.15) to induce false positives on MiraData-9k~\cite{mirabench}. 

As shown in Tab.~\ref{tab:ablate_vg}, under the noisier setting ($t$=0.15) with more false positives, \textit{SANTA} surpasses the existing SOTA hallucination mitigation method HACL~\cite{hacl} by a clear margin. Specifically, it achieves +2.1 and +0.6 absolute improvements on Avg. $\text{F1}_\text{Obj}$ and Avg. $\text{F1}_\text{Act}$ under HalFscore, and +2.9 and +1.1 under weighted-HalFscore, respectively. These results demonstrate that SANTA remains robust and effectively mitigates both object and action hallucinations, even when trained with imperfect object tracking results.


\section{Additional Qualitative Results}
\begin{figure}[t]
  \centering
  \includegraphics[width=0.5\textwidth, height=0.5\textheight, keepaspectratio]{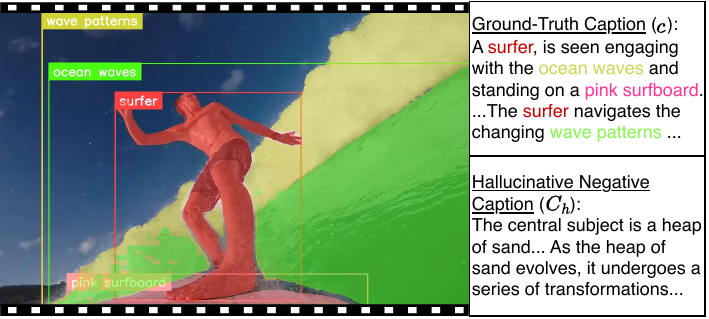}

  \caption{
    Qualitative results on hallucinative and object/action tracklets from MiraData-9k~\cite{mirabench}.
  }
  \label{fig:suppl_hallu_neg}

\end{figure}

\begin{figure*}[t!]
  \centering
  \includegraphics[width=1.0\textwidth]{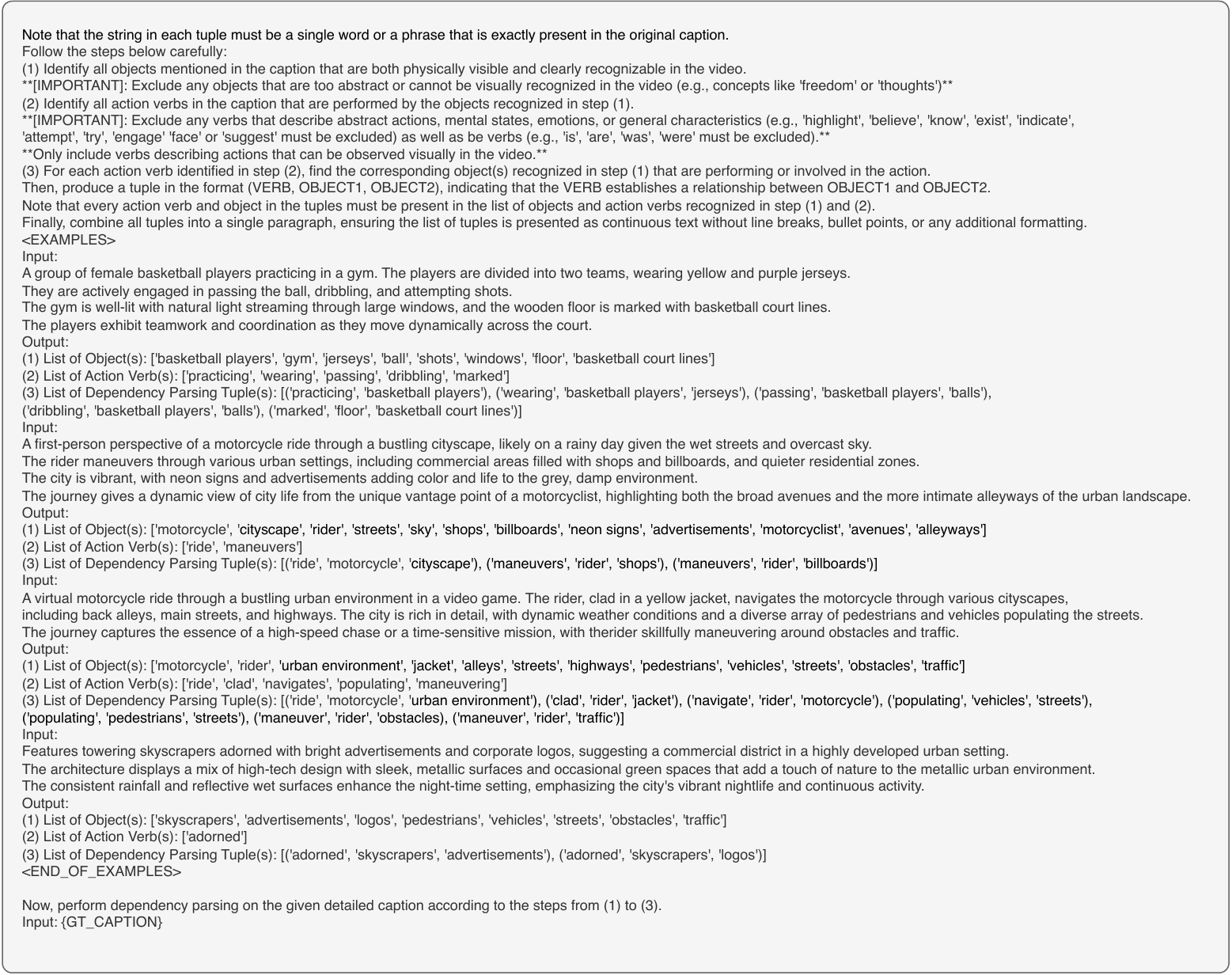}

  \caption{The prompting template used to prompt GPT-4o~\cite{gpt4o} with few-shot demonstrations to perform the parsing task. 
  }
  \label{fig:prompt_template}

\end{figure*}

\paragraph{Results on Hallucinative Negative and Object/Action Tracklets.}
Fig.~\ref{fig:suppl_hallu_neg} shows a surfing scene video frame, ground-truth caption $c$, and the hallucinative negative caption ($C_h$). In the visualized video frame, the presence of sea wave elements implies hallucination of ``heap of sand...'', etc. in $C_h$. The object tracklets are denoted as colored patches, and the action tracklets are derived from the associated two object tracklets (e.g., ``stand'' is associated with ``surfer'' and ``pink surfboard''). These relationships are determined by GPT4o~\cite{gpt4o} via prompting with the ground-truth caption, as described in Fig.~\ref{fig:prompt_template}.

\paragraph{Additional Results on MiraData-9k.}
In Fig.~\ref{fig:suppl_qualitative_mira}, we present additional qualitative results of applying \textit{SANTA} with LLaAV-Video~\cite{llavavideo} on MiraData-9k~\cite{mirabench}, which exhibits object and action faithfulness within its generated captions.

\begin{figure*}[t]
  \centering
  \includegraphics[width=3.5\textwidth, height=0.95\textheight, keepaspectratio]{figures/images/suppl_qualitative_mira.pdf}

  \caption{
    Qualitative comparison of video captions predicted by HACL~\cite{hacl} and \textit{SANTA} on MiraData-9k~\cite{mirabench}. Note that words highlighted in \textcolor{ForestGreen}{green} indicate action faithfulness, while those in \textcolor[HTML]{C23B22}{red} indicate action hallucination. Similarly, words in \textcolor{blue}{blue} represent object faithfulness, whereas those in \textcolor{orange}{orange} denote object hallucination.
  }
  \label{fig:suppl_qualitative_mira}

\end{figure*}

\end{document}